\documentclass{article}

\usepackage[final]{neurips_distshift_2022}

\usepackage[utf8]{inputenc} % allow utf-8 input
\usepackage[T1]{fontenc}    % use 8-bit T1 fonts
\usepackage{hyperref} 
\hypersetup{colorlinks=true,                
    breaklinks=true,                
    urlcolor= black,                
    linkcolor= blue,                
    bookmarksopen=false,
    filecolor=black,
    citecolor=blue,
}
\usepackage{url}            % simple URL typesetting
\usepackage{booktabs}       % professional-quality tables
\usepackage{amsfonts}       % blackboard math symbols
\usepackage{nicefrac}       % compact symbols for 1/2, etc.
\usepackage{microtype}      % microtypography
\usepackage{xcolor}         % colors

\newcommand{\Dd}[1]{\mathcal{D}^{#1}}

\newcommand{\LL}{\mathcal{L}}

\newcommand{\Ss}{\mathcal{S}}

\def\1{\bm{1}}

\DeclareMathAlphabet{\mathsfit}{\encodingdefault}{\sfdefault}{m}{sl}
\SetMathAlphabet{\mathsfit}{bold}{\encodingdefault}{\sfdefault}{bx}{n}

\usepackage{url}            % simple URL typesetting

\usepackage{graphicx}

\usepackage{float}
\usepackage{amsmath}
\usepackage{multirow}
\usepackage{csquotes}
\usepackage{textgreek}
\usepackage{todonotes}
\usepackage{rotating}
\usepackage{svg}
\usepackage{caption}
\usepackage{subcaption}
\usepackage{verbatim}
\newtheorem{definition}{Definition}
\title{Explanation Shift: Detecting distribution shifts on tabular data via the explanation space}

\author{%
  Carlos Mougan \\
  University of Southampton\\
  United Kingdom\\
  \texttt{C.Mougan@soton.ac.uk} \\
   \And
   Klaus Broelemann \\
   Schufa Holding AG\\
   Wiesbaden, Germany\\
   \texttt{klaus.broelemann@schufa.de} \\
   \And 
    Gjergji Kasneci\\
    University of Tuebingen\\
    Tuebingen, Germany\\
    \texttt{gjergji.kasneci@uni-tuebingen.de} \\
    \And 
    Thanassis Tiropanis\\
    University of Southampton\\
    United Kingdom\\
   \texttt{t.tiropanis@southampton.ac.uk} \\
     \And 
    Steffen Staab\\
    University of Stuttgart, Germany\\
    University of Southampton, UK\\
   \texttt{s.r.staab@soton.ac.uk} \\
}

\begin{document}

\maketitle

\begin{abstract}
As input data distributions evolve, the predictive performance of machine learning models tends to deteriorate. In the past, predictive performance was considered the key indicator to monitor. However, explanation aspects have come to attention within the last years. In this work, we investigate how model predictive performance and model explanation characteristics are affected under distribution shifts and how these key indicators are related to each other for tabular data.
We find that the modeling of explanation shifts can be a better indicator for the detection of predictive performance changes than state-of-the-art techniques based on representations of distribution shifts. We provide a mathematical analysis of different types of distribution shifts as well as synthetic experimental examples.
\end{abstract}

\section{Introduction}

Machine learning theory gives us the means to forecast the quality of ML models on unseen data, provided that this data is sampled from the same distribution as the data used to train and to evaluate the model. If unseen data is sampled from a different distribution, model quality may deteriorate.

Model monitoring tries to signal and possibly even quantify such decay of trained models. Such monitoring is challenging, because only in few applications unseen data comes with labels that allow for monitoring model quality directly. Much more often,  deployed ML models encounter unseen data for which target labels are lacking or biased~\cite{DBLP:conf/nips/RabanserGL19,DesignMLSystems}.   

%the deployment data has accessible labels, and predictive quality measures can be calculated. There are many use cases where there are no labels for the data used in deployment, or the labels are difficult to obtain.
%\steffen{Every time you say `performance' I think of run time, never of quality. Maybe this is just me. At least `performance' is ambiguous.}

%\todo{@Gjergji: Add some real-world examples here: Real-time advertisement ranking on the web or mobile apps (labels exist only partially and may be biased, i.e., the observed clicks), credit scoring, fraud detection (no labels after deployment; labels are only collected in batches and serve for the development of the next model). Also, state that most of these use cases are based on heterogeneous tabular data.}

%Hence, typically, the only data available in ML applications are labeled source data and unlabeled deployment data~\cite{garg2022leveraging}.
Detecting changes in the quality of deployed ML models in the absence of labeled data is a challenging question both in theory and practice~\cite{DBLP:conf/aaai/RamdasRPSW15,DBLP:conf/nips/RabanserGL19}. In practice, some of the most straightforward approaches
are based on statistical distances between training and unseen data distributions~\cite{continual_learning,clouderaff} or on actual model predictions~\cite{garg2022leveraging,garg2021ratt,mougan2022monitoring}. The shortcomings of these measures of \emph{distribution shifts} is that they do not relate changes of distributions to how they incur effects in the trained models.

% xAI
The field of explainable AI has emerged as a way to understand model decisions ~\cite{molnar2019} and interpret the inner workings of black box models~\cite{guidotti_survey}. The core idea of this paper is to use \emph{explanation shift} for signaling distribution shift that affect the model behavior. We newly define explanation shift to be constituted by the statistical comparison between how predictions from source data are explained and how predictions on unseen data are explained.  Explanation shift goes beyond the mere recognition of changes in data distributions towards the recognition of changes of how data distributions relate to the models' inner workings.

%As we will show, this idea even carries beyond predictive performance. While in the past  ML models' quality was mainly related to quality measures such as accuracy, f1 score, or mean squared error, awareness has increased in recent years that \enquote{even models developed with the best of intentions may exhibit discriminatory biases, perpetuate inequality, or perform less well for historically disadvantaged groups ~\cite{barocas-hardt-narayanan}}. Therefore, we introduce the notion \emph{decay of fairness} and compare how well methods based on distribution shift vs.\ methods based on explanation shift can deal with it.

We study the problem of detecting distribution changes that impact model predictive performance on tabular data, which still constitutes a major field of application for machine learning. In contrast, most recent research on model degradation proposes monitoring methods for modalities such as texts or images and suggests invariances on the latent spaces of deep neural models, which are hardly applicable to tabular data, for which likewise progress has mostly been lacking in recent years. In summary, our contributions are:
\begin{itemize}
    \item We propose measures of explanation shifts as a key indicator for detecting distribution shift that affect model behavior.
    
    \item We provide a mathematical analysis of three synthetic examples that shows how simple, but key types of distribution shift interact with linear models such that measures of explanation shift become much better indicators of model decay than measures of distribution shift or prediction shift. 
    
    %\item \textcolor{red}{We compare how measures of shifts of distributions, predictions, and explanations affect the task of quantifying model degradation in terms of predictive performance. }
    
    %\item \textcolor{red}{We provide comprehensive experimental evaluations and comparisons on real-world data concerned with  detecting and quantifying model decay.}
\end{itemize}

\section{Methodology}
\subsection{Formalization}

The objective of supervised learning is to induce a function $f_\theta:X \to Y$, where $f_\theta$ is from a family of functions $f_\theta \in F$, from training set  $\Dd{tr}=\{(x_0^{tr},y_0^{tr})\ldots, (x_n^{tr},y_n^{tr})\} \subseteq \mathcal{X} \times \mathcal{Y}$ where $\mathcal{X} \times \mathcal{Y}$ denote the domain of predictors $X$ and target $Y$ respectively. 
The estimated hypothesis $f_\theta$ is expected to generalize well on  novel, previously unseen data $\Dd{new}=\{x_0^{new}, x_1^{new}, \ldots \} \subseteq \mathcal{X}$, for which the target labels are unknown. The traditional machine learning assumption is that training data $\Dd{tr}$ and novel data  $\Dd{new}$ are sampled from the same underlying distribution $P(X,Y)$. If we have a hold-out test data set $\Dd{te}=\{(x_0^{te},y_0^{te})\ldots, (x_m^{te},y_m^{te})\} \subseteq \mathcal{X} \times \mathcal{Y}$ disjoint from $\Dd{tr}$, but also sampled from $P(X,Y)$, one may use $\Dd{te}$ to estimate performance indicators for $\Dd{new}$. Commonly, novel data is sampled from a  distribution $P'(X,Y)$ that is different from $P(X,Y)$.% and there is no hold-out data set sampled from $P'(X,Y)$ to estimate model performance. 
We use $\Dd{ood}\subseteq \mathcal{X} \times \mathcal{Y}$ to refer to such novel, out-of-distribution data.

\begin{definition}\textit{(Feature Attribution Explanation)}
We write $\mathcal{S}$ for an explanation function that takes a model $f$ with parameter $\theta$ and data of interest $x$ and returns the calculation of the Shapley values $\mathcal{S}(f_\theta, x)$, with $p$ being the exact dimensions of the predictor $x$ and signature $\mathcal{S}:F\times \mathcal{X}\to \mathbb{R}^p$
\end{definition}
%\steffen{What is the signature?}\carlos{can i say $dim(x_i) \equiv dim(\mathcal{S}(f_\theta,x_i))$} \klaus{You could write something like $\mathcal{S}:F\times \mathcal{X}\to \mathbb{R}^p$} \klaus[inline]{Might be better to define it for a single feature vector $x$. Batch computation is an implementation detail.}\klaus{Be nice to the reader. $S$ (the subsets) and $\mathcal{S}$ (the explanation function) look quite similar. Consider to use e.g. $T$ for the former one.}

\begin{definition}\textit{(Explanation Shift)}
For a measure of statistical distance between two explanations of the model $f_\theta$ between $X$ and $X'$, we write $d(\mathcal{S}(f_\theta, X),\mathcal{S}(f_\theta, X'))$ .
\end{definition}

\subsection{Explanation Shift: Detecting model performance changes via the explanation space}\label{subsec:explanationShiftMethods}

The following section provides three different examples of situations where changes in the explanation space can correctly account for model behavior changes. Where statistical checks on the input data (1) cannot detect changes, (2) require sophisticated methods to detect these changes, or (3) detect changes that do not affect model behavior. For simplicity the model used  in the analytical examples is a linear regression where, if the features are independent, the Shapley value can be estimated by $\Ss(f_\theta, x_i) = a_i(x_i-\mu_i)$, where $a_i$ are the coefficients of the linear model and $\mu_i$ the mean of the features \cite{DBLP:journals/corr/ShapTrueModelTrueData}. Moreover, in the experimental section, we provide examples with synthetic data, and non-linear models.

\subsubsection{Detecting multivariate shift}

One challenging type of distribution shift to detect is cases where the univariate distributions for each feature $j$ are equal between the source and the unseen dataset. Moreover, what changes are the distribution interdependencies among different features. Multiple univariate testing offers comparable performance to multivariate testing~\cite{DBLP:conf/nips/RabanserGL19}, but comparing distributions on high dimensional spaces is not an easy task. The following  examples aim to demonstrate that Shapley values account for covariate interaction changes while a univariate statistical test will provide false negatives. 

%\steffen{column vs row vectors don't match in the following}\carlos{check?}
\textbf{Example 1: \textit{Multivariate Shift}}\textit{
Let $X = (X_1,X_2) \sim  N\left(\begin{bmatrix}\mu_{1}  \\ \mu_{2} \end{bmatrix},\begin{bmatrix}\sigma^2_{x_1} & 0 \\0 & \sigma^2_{x_2} \end{bmatrix}\right)$
and $X^{ood} = (X^{ood}_1,X^{ood}_2) \sim  N\left(\begin{bmatrix}\mu_{1}  \\ \mu_{2} \end{bmatrix},\begin{bmatrix} \sigma^2_{x_1} & \rho\sigma_{x_1}\sigma_{x_2}  \\ \rho\sigma_{x_1}\sigma_{x_2} & \sigma^2_{x_2}\end{bmatrix}\right)$. We fit a linear model 
$f_\theta(X_1,X_2) = \gamma + a\cdot X_1 + b \cdot X_2.\hspace{0.5cm}$  $X_1$ and $X_2$ are identically distributed with $X_1^{ood}$ and $X_2^{ood}$, respectively, while this does not hold for the corresponding SHAP values $\Ss_j(f_\theta,X)$ and $\Ss_j(f_\theta,X^{ood})$. Analytical demonstration in the Appendix}

%The above theorem works under the assumption of linear regression and that the covariate term $\rho=1$, is a not-so-common situation in ML applications.

\subsubsection{Detecting posterior distribution shift}
One of the most challenging types of distribution shift to detect are cases where distributions are equal between source and unseen data-set $P(X^{tr}) = P(X^{ood})$ and the target variable  $P(Y^{tr}) = P(Y^{ood})$ and what changes are the relationships that features have with the target $P(Y^{tr}|X^{tr}|) \neq  P(Y^{ood}|X^{ood}|)$, this kind of distribution shift is also known as concept drift or posterior shift~\cite{DesignMLSystems} and is especially difficult to notice, as it requires labeled data to detect. The following example compares how the explanations change for two models fed with the same input data and different target relations.

\textbf{Example 2: \textit{Posterior shift}}\textit{
Let $X = (X_1,X_2) \sim N(\mu,I)$, and $X^{ood}= (X^{ood}_1,X^{ood}_2) \sim N(\mu,I)$, where $I$ is an identity matrix of order two and $\mu = (\mu_1,\mu_2)$. We now create two synthetic targets $Y=a + \alpha \cdot X_1 + \beta \cdot X_2 + \epsilon$ and $Y^{ood}=a + \beta \cdot X_1 + \alpha \cdot X_2 + \epsilon$. Let $f_\theta$ be a linear regression model trained on $f(X,Y)$ and $g_\psi$ another linear model trained on $(X^{ood},Y^{ood})$. Then $P(f_\theta(X)) = P(g_\psi(X^{ood}))$, $P(X) = P(X^{ood})$ but $\Ss(f_\theta,X)\neq \Ss(g_\psi, X)$}. 

\subsubsection{Shifts on uninformative features by the model}

Another typical problem is false positives when a statistical test flags a distribution difference between source and unseen distributions that do not affect the model behavior\cite{grinsztajn:hal-03723551}. One of the intrinsic properties that Shapley values satisfy is the \enquote{Dummy}, where a feature $j$ that does not change the predicted value, regardless of which coalition the feature is added, should have a Shapley value of $0$. If $\mathrm{val}(S\cup \{j\}) = \mathrm{val}(S)$ for all $S\subseteq \{1,\ldots, p\}$ then $\Ss(f_\theta, x_j)=0$.

\textbf{Example 3: \textit{Unused features}}\textit{
Let $X = (X_1,X_2,X_3) \sim N(\mu,c\cdot I)$, and $X^{ood}= (X^{ood}_1,X^{ood}_2,X^{ood}_3) \sim N(\mu,c'\cdot I)$, where $I$ is an identity matrix of order three and $\mu = (\mu_1,\mu_2,\mu_3)$. We now create a synthetic target $Y=a_0 + a_1 \cdot X_1 + a_2 \cdot X_2 + \epsilon$ that is independent of $X_3$. We train a linear regression $f_\theta$ on $(X,Y)$, with coefficients $a_0,a_1,a_2,a_3$. Then $P(X_3)$ can be different from $P(X_3^{ood})$ but $\Ss_3(f_\theta, X) = \Ss_3(f_\theta,X^{ood})$}

\section{Experiments}\label{sec:experiments}

The experimental section explores the detection of distribution shift on synthetic examples. We perform statistical testing between input data distributions and explanation space.

\subsection{Detecting multivariate shift}

Given two bivariate normal distributions $X = (X_1,X_2) \sim  N\left(0,\begin{bmatrix}1 & 0 \\0& 1 \end{bmatrix}\right)$ and $X^{ood} = (X^{ood}_1,X^{ood}_2) \sim  N \left( 0,\begin{bmatrix}1 & 0.2 \\0.2 & 1 \end{bmatrix}\right)$, then, for each feature $j$ the underlying distribution is equally distributed between $X$ and $X^{ood}$, $\forall j \in \{1,2\}: P(X_j) = P(X^{ood}_j)$, and what is different are the interaction terms between them. We now create a synthetic target $Y=X_1\cdot X_2 + \epsilon$ with $\epsilon \sim N(0,0.1)$ and fit a gradient boosting decision tree  $f_\theta(X)$. Then we compute the SHAP explanation values for $\mathcal{S}(f_\theta,X)$ and $\mathcal{S}(f_\theta,X^{ood})$

\begin{table}[ht]
\centering
\caption{Displayed results are the one-tailed p-values of the Kolmogorov-Smirnov test comparison between two underlying distributions. Small p-values indicates that compared distributions would be very unlikely  to be equally distributed. SHAP values correctly indicate the interaction changes that individual distribution comparisons cannot detect}\label{table:multivariate}
\begin{tabular}{c|cc}
Comparison                                 & \textbf{p-value} & \textbf{Conclusions} \\ \hline
$P(X_1)$, $P(X^{ood}_1)$                        & 0.33                        & Not Distinct                         \\
$P(X_2)$, $P(X^{ood}_2)$                        & 0.60                        & Not Distinct                          \\
$\Ss_1(f_\theta,X)$, $\Ss_1(f_\theta,X^{ood})$ & $3.9\mathrm{e}{-153}$        & Distinct                              \\
$\Ss_2(f_\theta,X)$, $\Ss_2(f_\theta,X^{ood})$ & $2.9\mathrm{e}{-148}$        & Distinct   
\end{tabular}
\end{table}

Having drawn $50,000$ samples from both $X$ and $X^{ood}$, in Table~\ref{table:multivariate}, we evaluate whether changes on the input data distribution or on the explanations are able to detect changes on covariate distribution.
%\steffen{... complete... What are your hypotheses/null-hypotheses? You jump over too many steps, which makes your proposition become awkwardly ambiguous.}
For this, we compare the one-tailed p-values of the Kolmogorov-Smirnov test between the input data distribution, and the explanations space.  Explanation shift correctly detects the multivariate distribution change that univariate statistical testing can not detect.

\subsubsection{Detecting posterior shift}

Given a bivariate normal distribution  $X = (X_1,X_2) \sim  N(1,I)$ where $I$ is an identity matrix of order two. We now create two synthetic targets $Y= X_1^2 \cdot X_2 + \epsilon$ and $Y^{ood}=X_1 \cdot X_2^2 + \epsilon$ and fit two machine learning models $f_\theta$ on $(X,Y)$ and $h_\Upsilon$ on $(X,Y^{ood})$. Now we compute the SHAP values for $\mathcal{S}(f_\theta,X)$ and $\mathcal{S}(h_\Upsilon,X)$
%\steffen{If I understand it correctly, the use of $f$ and $g$ is here completely wrong. First, you train a fuction $f_\theta(X)=Y$ (and not $f_\theta(X,Y)$), second, $g$ was previously used to run on distances, now it would run on X --- don't do such a sudden change of semantics for your symbols. Rather say: $f_\phi(X)=Y^{ood}$. Table 2 needs update, too. }\carlos{Yes, you are right, its not G, but another model $h_\Upsilon$}

\begin{table}[ht]
\centering
\caption{Distribution comparison for synthetic posterior shift. Displayed results are the one-tailed p-values of the Kolmogorov-Smirnov test comparison between two underlying distributions }\label{table:posterior}
\begin{tabular}{c|c}
Comparison                                              & \textbf{Conclusions} \\ \hline
$P(X)$, $P(X^{ood})$                                    & Not Distinct         \\
$P(Y)$, $P(Y^{ood})$                                    & Not Distinct         \\
$P(f_\theta(X))$, $P(h_\Upsilon(X^{ood}))$                  & Not Distinct         \\
$\Ss(f_\theta,X)$, $\Ss(h_\Upsilon,X)$                    & Distinct             \\
\end{tabular}
\end{table}

In Table~\ref{table:posterior}, we see how the distribution shifts are not able to capture the change in the model behavior while the SHAP values are different. The \enquote{Distinct/Not distinct} conclusion is based on the one-tailed p-value of the Kolmogorov-Smirnov test with a $0.05$ threshold drawn out of $50,000$ samples for both distributions. As in the theoretical example, in table \ref{table:posterior} SHAP values can detect a relational change between $X$ and $Y$, even if both distributions remain equivalent.

\section{Conclusions}
Traditionally, the problem of detecting model degradation has relied on measurements of shifting input data distributions or shifting distributions of predictions. In this paper, we have provided theoretical and experimental evidence that explanation shift can be a more suitable indicator to detect and quantify decay of predictive performance. We have provided mathematical analysis examples and synthetic data experimental evaluation. We found that measures of explanation shift can outperform measures of distribution and prediction shift. 

\textbf{Limitations:} Without any assumptions on the type of shift, estimating model decay is a challenging task, where no estimator will be the best under all the types of shift \cite{garg2022leveraging}. We compared how well measures of explanation shift would perform relative to distribution shift and found encouraging results. The  potential utility of explanation shifts as indicators for predictive performance and fairness in computer vision or natural language processing tasks remains an open question. We have used Shapley values to derive indications of explanation shifts, but we believe that other AI explanation techniques and specifically, other feature attribution methods, logical reasoning, argumentation,  or counterfactual explanations, may be applicable and come with their own advantages.

%As explainability technique we have used Shapley values, there is further explainable AI techniques such as more feature attribution methods or contrafactual explanations that can be applied. 

\subsection*{Reproducibility Statement}\label{sec:reproducibility}
To ensure reproducibility, we make the data, code repositories, and experiments publicly available
\footnote{\url{https://anonymous.4open.science/r/ExplanationShift-691E}}.
%Also, an open source Python package is released with the methods used (to be released upon acceptance). 
For our experiments, we used default \texttt{scikit-learn} parameters \cite{pedregosa2011scikit}. We describe the system requirements and software dependencies of our experiments. Experiments were run on a 16 vCPU server with 60 GB RAM.

\subsection*{Acknowledgments}
This work has received funding by the European Union’s Horizon 2020 research and innovation programme under the Marie
Skłodowska-Curie Actions (grant agreement number 860630) for
the project : \enquote{NoBIAS - Artificial Intelligence without Bias}. Furthermore, this work reflects only the authors’ view and the European Research Executive Agency (REA) is not responsible for any
use that may be made of the information it contains.

\bibliographystyle{apalike}
\bibliography{references}

\appendix
\section{Foundations and Related Work}\label{sec:foundations}

\subsection{Explainable AI}
Explainability has become an important concept in legal and ethical guidelines for data processing, and machine learning applications ~\cite{selbst2018intuitive}. A wide variety of methods have been developed aiming to account for the decision of algorithmic systems ~\cite{guidotti_survey,DBLP:conf/fat/MittelstadtRW19,DBLP:journals/inffus/ArrietaRSBTBGGM20}. One of the most popular approaches to machine learning explainability has been the use of Shapley values to attribute relevance to features used by the model~\cite{shapTree,lundberg2017unified}. The Shapley value is a concept from coalition game theory that aims to allocate the surplus generated by the grand coalition in a game to each of its players~\cite{shapley}. In a general sense, the Shapley value $\mathcal{S}_j$ for the $j$'th player can be defined via a value function $\mathrm{val}:2^N \to \mathbb{R}$ of players in $T$:

\begin{gather}
\mathcal{S}_j(\mathrm{val}) = \sum_{T\subseteq N\setminus \{j\}} \frac{|T|!(p-|T|-1)!}{p!}(\mathrm{val}(T\cup \{j\}) - \mathrm{val}(T))
\end{gather}

In the case of features, $T$ is a subset of $N=\{1,\ldots,p\}$, i.e., the features used in the model, and $x$ is the vector of feature values of the instance to be explained. The term $\mathrm{val}_x(T)$ represents the prediction for the feature values in $T$ that are marginalized over features that are not included in $T$:
\begin{gather}
\mathrm{val}_x(T) = E_{X_{N\setminus T}}[\hat{f}(X)|X_T=x_T]-E_X[\hat{f}(X)]
\end{gather}

%\steffen{The first expected value does not make sense as the index in the E does have a different dimensionality than the X. And this seems to be only one bug}\klaus[inline]{I still think this equation is valid: we condition on the features in $T$, so the expectation has to be done overall feature, not in $T$. So $E_{X_{N\setminus T}}$ is the expectation we want to build. If we ignore the ordering, we could write $\hat{f}(X_T,X_{N\setminus T})$ instead of $\hat{f}(X)$, which might make it clearer, but formally less correct.}

It is important to differentiate between the theoretical Shapley values and the different implementations that approximate them. We use  TreeSHAP as an efficient implementation of an approach for tree-based models of Shapley values~\cite{shapTree,molnar2019}, particularly we use the observational (or path-dependent) estimation  ~\cite{DBLP:journals/corr/abs-2207-07605,DBLP:conf/nips/FryeRF20,DBLP:journals/corr/ShapTrueModelTrueData} and for linear models we use the correlation dependent implementation that takes into account feature dependencies \cite{DBLP:journals/ai/AasJL21}.

\subsection{Related Work}
Evaluating how two distributions differ has been a widely studied topic in the statistics and statistical learning literature~\cite{statisticallearning,datasetShift}.~\cite{DBLP:conf/nips/RabanserGL19} provide a comprehensive empirical investigation, examining how
dimensionality reduction and two-sample testing might be combined to produce a practical pipeline for detecting distribution shifts in real-life machine learning systems. A few popular techniques to detect out-of-distribution data using neural networks are based on the prediction space~\cite{fort2021exploring,NEURIPS2020_219e0524}, using the maximum softmax probabilities/likelihood as a confidence score, extracting information out of the gradient space~\cite{DBLP:journals/corr/GradientShift}, fitting a Gaussian distribution to the embedding or using the Mahalanobis distance for out-of-distribution detection~\cite{DBLP:conf/iclr/HendrycksG17,DBLP:conf/nips/LeeLLS18,DBLP:journals/corr/reliableShift}. These methods are built explicitly for neural networks, and often they can not be directly applied to traditional machine learning techniques. In our work, we are focusing specifically on tabular data where techniques such as gradient boosting decision trees achieve state of the art on model performance~\cite{grinsztajn:hal-03723551,DBLP:journals/corr/abs-2101-02118,BorisovNNtabular}. 

% Shap original paper suggestion
The first approach of using explainability to detect changes in the model was suggested by~\cite{shapTree} who monitored the SHAP value contribution in order to identify possible bugs in the pipeline. This technique was initially used to account for previously unnoticed bugs in a local monitoring scenario in the machine learning production pipeline. In our work, we study how SHAP value changes can be used as an indicator to monitor prediction performance and fairness.

\section{Analytical examples}
This section covers the analytical examples demonstrations presented in the Section \ref{subsec:explanationShiftMethods} of the main body of the paper.
\subsection{Multivariate shift}
\textbf{Example 1: \textit{Multivariate Shift}}\textit{
Let $X = (X_1,X_2) \sim  N\left(\begin{bmatrix}\mu_{1}  \\ \mu_{2} \end{bmatrix},\begin{bmatrix}\sigma^2_{x_1} & 0 \\0 & \sigma^2_{x_2} \end{bmatrix}\right)$
and $X^{ood} = (X^{ood}_1,X^{ood}_2) \sim  N\left(\begin{bmatrix}\mu_{1}  \\ \mu_{2} \end{bmatrix},\begin{bmatrix} \sigma^2_{x_1} & \rho\sigma_{x_1}\sigma_{x_2}  \\ \rho\sigma_{x_1}\sigma_{x_2} & \sigma^2_{x_2}\end{bmatrix}\right)$. We fit a linear model 
$f_\theta(X_1,X_2) = \gamma + a\cdot X_1 + b \cdot X_2.\hspace{0.5cm}$  $X_1$ and $X_2$ are identically distributed with $X_1^{ood}$ and $X_2^{ood}$, respectively, while this does not hold for the corresponding SHAP values $\Ss_j(f_\theta,X)$ and $\Ss_j(f_\theta,X^{ood})$.}
\begin{gather}
\Ss_1(f_\theta,x) = a(x_1 - \mu_1)\\
\Ss_1(f_\theta,x^{ood}) =\\
=\frac{1}{2}[\mathrm{val}(\{1,2\}) - \mathrm{val}(\{2\})] + \frac{1}{2}[\mathrm{val}(\{1\}) - \mathrm{val}(\emptyset)] \\
\mathrm{val}(\{1,2\}) = E[f_\theta|X_1=x_1, X_2=x_2] = a x_1 + b x_2\\
\mathrm{val}(\emptyset) = E[f_\theta]= a \mu_1 + b  \mu_2 \\
\mathrm{val}(\{1\}) = E[f_\theta(x) | X_1 = x_1] +b\mu_2 \\
\mathrm{val}(\{1\}) = \mu_1 +\rho \frac{\rho_{x_1}}{\sigma_{x_2}}(x_1-\sigma_1)+b \mu_2\\
\mathrm{val}(\{2\}) = \mu_2 +\rho \frac{\sigma_{x_2}}{\sigma_{x_1}}(x_2-\mu_2)+a\mu_1 \\
\rightarrow \Ss_1(f_\theta,x^{ood})\neq a(x_1 - \mu_1)
\end{gather}

\subsection{Posterior Shift}
\textbf{Example 2: \textit{Posterior shift}}\textit{
Let $X = (X_1,X_2) \sim N(\mu,I)$, and $X^{ood}= (X^{ood}_1,X^{ood}_2) \sim N(\mu,I)$, where $I$ is an identity matrix of order two and $\mu = (\mu_1,\mu_2)$. We now create two synthetic targets $Y=a + \alpha \cdot X_1 + \beta \cdot X_2 + \epsilon$ and $Y^{ood}=a + \beta \cdot X_1 + \alpha \cdot X_2 + \epsilon$. Let $f_\theta$ be a linear regression model trained on $f(X,Y)$ and $g_\psi$ another linear model trained on $(X^{ood},Y^{ood})$. Then $P(f_\theta(X)) = P(g_\psi(X^{ood}))$, $P(X) = P(X^{ood})$ but $\Ss(f_\theta,X)\neq \Ss(g_\psi, X)$}. 
\begin{gather}
X  \sim N(\mu,\sigma^2\cdot I), X^{ood}\sim N(\mu,\sigma^2\cdot I)\\
\rightarrow P(X) = P(X^{ood})\\
Y \sim a + \alpha N(\mu, \sigma^2) + \beta N(\mu, \sigma^2) + N(0, \sigma^{'2})\\
Y^{ood} \sim a + \beta N(\mu, \sigma^2) + \alpha N(\mu, \sigma^2) + N(0, \sigma^{'2})\\
\rightarrow P(Y) = P(Y^{ood})\\
\Ss(f_\theta,X) = \left( \begin{matrix}\alpha(X_1 - \mu_1)  \\\beta(X_2-\mu_2) \end{matrix}\right) \sim \left(\begin{matrix}N(\mu_1,\alpha^2 \sigma^2)  \\N(\mu_2,\beta^2 \sigma^2) \end{matrix}\right)\\
\Ss(g_\psi,X) =  \left( \begin{matrix}\beta(X_1 - \mu_1)  \\\alpha(X_2-\mu_2) \end{matrix}\right)\sim \left(\begin{matrix}N(\mu_1,\beta^2 \sigma^2)  \\N(\mu_2,\alpha^2 \sigma^2) \end{matrix}\right)\\
\mathrm{If} \quad \alpha \neq \beta \rightarrow \Ss(f_\theta,X)\neq \Ss(g_\psi,X)
\end{gather}
\subsection{Uninformative Features}

\textbf{Example 3: \textit{Unused features}}\textit{
Let $X = (X_1,X_2,X_3) \sim N(\mu,c\cdot I)$, and $X^{ood}= (X^{ood}_1,X^{ood}_2,X^{ood}_3) \sim N(\mu,c'\cdot I)$, where $I$ is an identity matrix of order three and $\mu = (\mu_1,\mu_2,\mu_3)$. We now create a synthetic target $Y=a_0 + a_1 \cdot X_1 + a_2 \cdot X_2 + \epsilon$ that is independent of $X_3$. We train a linear regression $f_\theta$ on $(X,Y)$, with coefficients $a_0,a_1,a_2,a_3$. Then $P(X_3)$ can be different from $P(X_3^{ood})$ but $\Ss_3(f_\theta, X) = \Ss_3(f_\theta,X^{ood})$}
\begin{gather}
X_3\sim N(\mu_3,c_3),X_3^{ood} \sim N(\mu_3^{'}, c_3^{'})\\
\mathrm{If} \quad  \mu_3^{'}\neq \mu_3 \quad\mathrm{or} \quad c_3^{'}\neq c_3 \rightarrow P(X_3)\neq P(X_3^{ood})\\
\Ss(f_\theta,X) = \left(\begin{bmatrix} a_1(X_1 - \mu_1)  \\a_2(X_2 - \mu_2)  \\a_3(X_3 - \mu_3)   \end{bmatrix} \right) = \left(\begin{bmatrix} a_1(X_1 - \mu_1)  \\a_2(X_2 - \mu_2)  \\0   \end{bmatrix} \right)\\
\Ss_3(f_\theta,X)= \Ss_3(f_\theta, X^{ood})
\end{gather}

\section{Synthetic data experiments}
This section covers the last experiment of uninformative features on synthetic data that aims at providing empirical evidence about using the explanation space as (cf. Section \ref{sec:experiments})  
\subsection{Uninformative features on synthetic data}

To have an applied use case of the theoretical example from the methodology section, we create a three-variate normal distribution $X = (X_1,X_2,X_3) \sim N(0,I_3)$, where $I_3$ is an identity matrix of order three. The target variable is generated  $Y=X_1\cdot X_2 + \epsilon$ being independent of $X_3$. For both, training and test data, $50,000$ samples are drawn. Then out-of-distribution data is created by shifting $X_3$, which is independent of the target, on test data $X^{ood}_3= X^{te}_3+1$.

\begin{table}[ht]
\centering
\caption{Distribution comparison when modifying a random noise variable on test data. Where $\LL$ is a metric evaluating the model predictive performance such as accuracy}\label{table:unused}
\begin{tabular}{c|c}
Comparison                                              & \textbf{Conclusions} \\ \hline
$P(X^{te}_3)$, $P(X^{ood}_3)$                                       & Distinct                \\
$\LL(f_\theta,X^{te})$, $\LL(f_\theta,X^{ood})$                     & Not Distinct            \\
$\Ss(f_\theta,X^{te})$, $\Ss(f_\theta,X^{ood})$                    & Not Distinct            \\
\end{tabular}
\end{table}

In Table~\ref{table:unused}, we see how an unused feature has changed the input distribution, but the explanation space and performance evaluation metrics remain the same. The \enquote{Distinct/Not Distinct} conclusion is based on the one-tailed p-value of the Kolmogorov-Smirnov test drawn out of $50,000$ samples for both distributions.

\end{document}